%% file: Paper-3963.tex
\documentclass[runningheads]{llncs}
\usepackage[T1]{fontenc}
\usepackage{graphicx}
\usepackage{hyperref}
\usepackage{mathtools}
\usepackage{amsmath}
\usepackage{amssymb}
\usepackage{multirow}
\usepackage{enumitem}
\usepackage{array}
\usepackage{caption}
\usepackage{subcaption}
\usepackage{float}
\usepackage{booktabs}
\usepackage{color}

\begin{document}
\title{Active Label Refinement for Robust Training of Imbalanced Medical Image Classification Tasks in the Presence of High Label Noise}
\titlerunning{Active label cleaning for imbalanced medical image classification}
\author{Bidur Khanal \inst{1}\and
Tianhong Dai\inst{3}\and
Binod Bhattarai\inst{3} \textsuperscript{*} \and
Cristian Linte\inst{1,2} \textsuperscript{*} }
%

\institute{ Center for Imaging Science, Rochester Institute of Technology, Rochester, NY, USA \and Biomedical Engineering, Rochester Institute of Technology, Rochester, NY, USA \and University of Aberdeen, Aberdeen, UK\\}

\authorrunning{B. Khanal {\it et al.}}

\renewcommand{\thefootnote}{*}
\footnotetext[1]{These authors share equal senior authorship.}
\renewcommand{\thefootnote}{\arabic{footnote}} 

\maketitle              

\begin{abstract}

The robustness of supervised deep learning-based medical image classification is significantly undermined by label noise in the training data. Although several methods have been proposed to enhance classification performance in the presence of noisy labels, they face some challenges: 1) a struggle with class-imbalanced datasets, leading to the frequent overlooking of minority classes as noisy samples; 2) a singular focus on maximizing performance using noisy datasets, without incorporating experts-in-the-loop for actively cleaning the noisy labels. To mitigate these challenges, we propose a two-phase approach that combines Learning with Noisy Labels (LNL) and active learning. This approach not only improves the robustness of medical image classification in the presence of noisy labels but also iteratively improves the quality of the dataset by relabeling the important incorrect labels, under a limited annotation budget. Furthermore, we introduce a novel Variance of Gradients approach in the LNL phase, which complements the loss-based sample selection by also sampling under-represented examples. Using two imbalanced noisy medical classification datasets, we demonstrate that our proposed technique is superior to its predecessors at handling class imbalance by not misidentifying clean samples from minority classes as mostly noisy samples. Code available at: \href{https://github.com/Bidur-Khanal/imbalanced-medical-active-label-cleaning.git}{https://github.com/Bidur-Khanal/imbalanced-medical-active-label-cleaning.git}

\end{abstract}
\section{Introduction}

Label noise poses a significant hurdle in the robust training of classifiers for medical image datasets, as it can distort the supervised learning process and compromise generalizability \cite{karimi2020deep,khanal2023investigating}. In real-world scenarios, factors such as the lack of quality annotation \cite{orting2020survey}, the use of NLP algorithms to extract labels from test reports \cite{irvin2019chexpert}, and the reliance on pseudo labels \cite{kuznetsova2020open} lead to high label noise in datasets. The impact of label noise is particularly severe in imbalanced medical datasets, where the class distribution is skewed \cite{li2023learning}. In recent years, numerous approaches, collectively referred to as Learning with Noisy Labels, have been proposed to train classifiers robustly in the presence of noisy labels \cite{han2018co,liu2021co,khanal2023improving}. These methods often employ a sample selection strategy based on the big-loss hypothesis, which suggests that samples with low incurred loss are likely to be clean, to distinguish clean samples from noisy ones. However, this simple hypothesis alone fails with highly imbalanced datasets, where minority or hard samples are mistakenly interpreted as noisy, necessitating a robust alternative.

Medical datasets are often highly imbalanced due to the varying prevalence of conditions or diseases, with some being rarer than others. For example, Dermatofibroma occurs less frequently than other skin conditions, so it is often under-represented in the dataset. Although there have been attempts to enhance robustness against noisy labels in imbalanced datasets \cite{xue2022robust,li2023learning}, the performance still falls short of that achieved without noisy labels. Furthermore, a model trained solely on noisy labels is unlikely to be trusted for medical inference and has limited potential for improvement unless data is cleaned. Therefore, establishing a two-step mechanism to initially train optimally on a noisy, imbalanced dataset and then progressively correct labels over time to improve performance is crucial. 

One such strategy involves incorporating the experts-in-the-loop to selectively relabel important noisy samples within a limited annotation budget over time. This approach is akin to active learning, which aims to label the most important examples from a pool of unlabeled samples to maximize task performance \cite{budd2021survey}. However, active learning assumes the existence of some initial accurately labeled data to train the first model, which is unfeasible with a noisy dataset, making it likely to fail without such a set from the start. A practical method should optimally train with the noisy dataset and then learn to identify the samples that need relabeling based on noise statistics. Several machine learning papers have proposed methods to actively clean labels \cite{lin2016re,zeni2019fixing,goh2023activelab}. Bernhardt {\it et al.} \cite{bernhardt2022active} proposed an active label-cleaning method for noisy Chest X-ray datasets with multiple annotators. However, this method, specifically designed for multi-annotator scenarios, struggles with highly imbalanced datasets.

In this work, we propose an approach to address two key challenges: robustly training a classifier on a noisy, imbalanced dataset and gradually cleaning important noisy samples by incorporating experts-in-the-loop to enhance classifier performance. For robust training, we modified the loss-based sample selection strategy used in LNL, which separates clean and noisy samples based on sample loss \cite{han2018co,liu2021co,li2020dividemix}, by incorporating a Variance of Gradients-based selection. The loss-based approach alone struggles with imbalanced datasets, as underrepresented samples often exhibit high loss values, leading to their mis-selection as noisy. The Variance of Gradients-based selection is robust against such bias and complements the loss-based method, compensating for the potential exclusion of underrepresented samples.

To summarize our contributions: (1) \textit{we propose an active label cleaning pipeline to iteratively clean noisy labels for medical image classification by incorporating experts-in-the-loop under a limited annotation budget}; (2) \textit{we introduce a novel Variance of Gradients-based example selection strategy to complement loss-based clean label selection, aiming to better handle highly underrepresented samples in highly imbalanced datasets with high label noise}; (3) \textit{we demonstrate that our proposed method outperforms its predecessor baseline methods, while limiting the annotation budget, as shown using both the imbalanced ISIC-2019 and long-tailed NCT-CRC-HE-100K datasets}.

\begin{figure}[h!]
\centering
\includegraphics[width=1\linewidth]{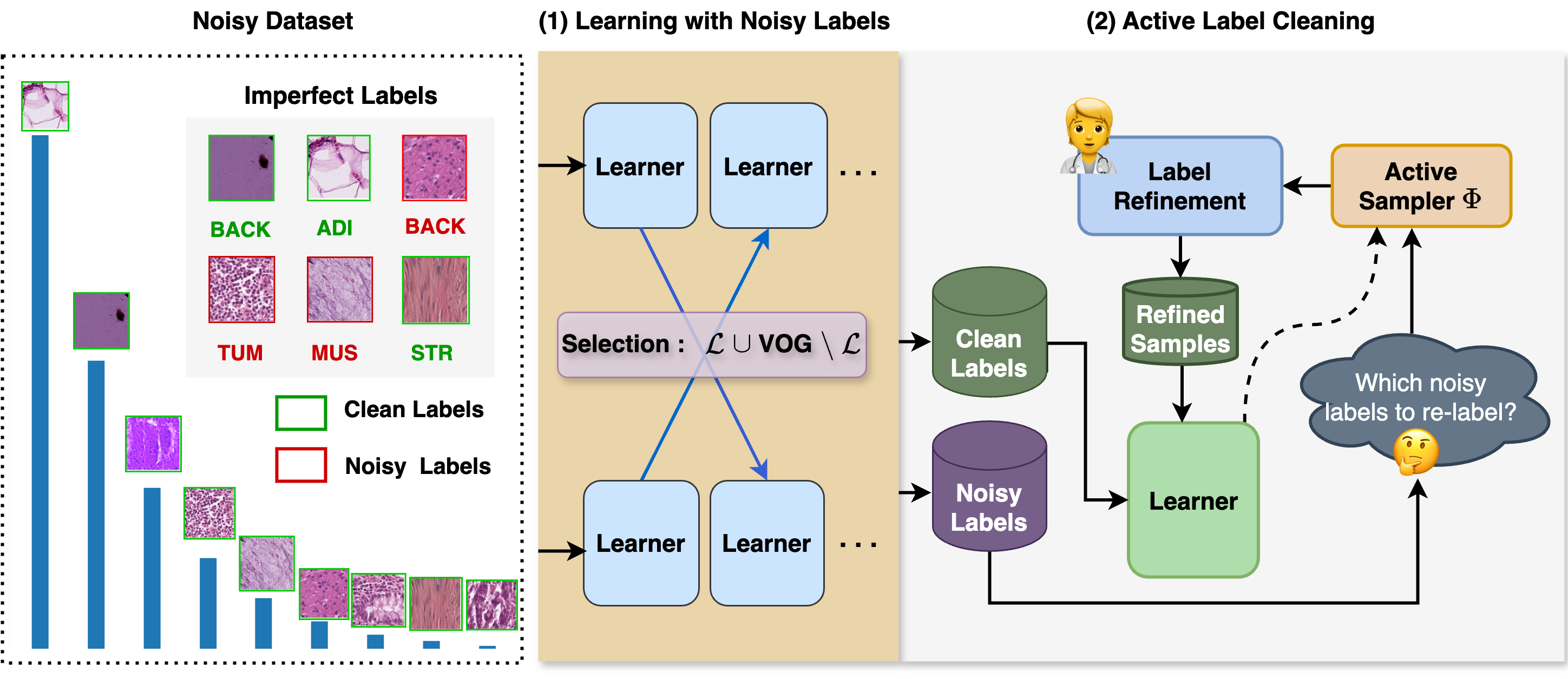}
\caption{Active Label Cleaning Pipeline: 1) Learning with Noisy Labels (LNL), where the clean-noisy selection process includes selections from both small Variance of Gradient (VOG) and small loss ($\mathcal{L}$) criteria; 2) Active Label Cleaning, wherein the noisy samples discarded by LNL are iteratively sampled using an active sampler ($\mathbf{\Phi}$) and relabeled.}
\label{fig:pipeline}
\end{figure}

\section{Methods}

\subsection{Label Noise Injection}
\label{label_noise}

We denote an imbalanced classification dataset as $\{(\mathbf{x}_i, y_i)\}_{i=1}^{N}$, where $N$ is the total number of instances, $x_i $ is an input and $y_i \in \{1,2,3,...,C\}$ is the corrresponding true label. Noisy dataset is created by injecting label noise by randomly flipping the true label $y_i$ with another class label $\hat{y_{i}}$ by a probability $p$ (i.e, noise rate), such that $\hat{y_{i}} \overset{p}{\sim} \{1,2,3,...,C\}\setminus\{y_{i}\}$.

\subsection{Overall Pipeline}

Overall, this is a two-stage pipeline as described in Fig. \ref{fig:pipeline}. In the first stage, we apply LNL to robustly train the model in the presence of noisy labels while concurrently identifying clean samples from the dataset. In the second stage, we use these clean samples to train a new model. Meanwhile, the remaining samples are ranked by an active sampler based on their importance using a ranking function. After ranking, the top $a_l$ samples are cleaned in each annotation round and used to train the model again until the total annotation budget $A_l$ is exhausted.

\subsection{LNL using Variance of Gradient}

In this first phase, we robustly train our model using an LNL method. The typical loss-based sample selection used in LNL to separate noisy samples from clean samples does not handle underrepresented samples in imbalanced datasets well, often misidentifying them as noisy samples.

We used a novel approach to regularize sample selection by using the Variance of Gradients (VOG) \cite{agarwal2022estimating}, instead of relying solely on loss-based selection. Similar to loss, VOG can also separate samples with clean labels from noisy ones. Unlike loss-based selection, VOG estimates the change in gradients over epochs rather than making selections based on statistics from a single epoch, thereby avoiding potential bias. The original paper \cite{agarwal2022estimating} computes the VOG of each sample at the image level and averages all the pixels to obtain a scalar value. This approach is unscalable as the dataset size and input-image resolution increase. Following \cite{shin2023loss}, we compute the VOG at the feature level, which significantly reduces the memory footprint (example: from 256$\times$256$\times N$ gradients to 512$\times N$ gradients, where 512 is the dimension of the ResNet18 feature representation).

Mathematically, let's assume $S_{ij} \in \mathbb{R}^D$ is the gradient vector ($ S_{ij} = \frac{\partial \mathcal{A}^{l}_{y_i}}{\partial x_i}$), for a sample $x_{i}$ at an epoch $j$, where $i \in \{1,2,..N\}$, $j \in \{1,2,..E\}$, and $\mathcal{A}^l_{y_i}$ is the class activation w.r.t given label $y_i$. Here, $N$ and $E$ represent the number of data samples and the number of epochs, respectively, while $D$ is the dimension of the gradient vector, equal to the feature dimension. Each sample $x_i$ has a gradient vector computed at various epochs, i.e. $\{S_{i1}, S_{i2},...S_{iE}\}$. The Variance of Gradient (VOG) of \(x_{i}\), at an epoch $j$, is given by:
\begin{equation}
VOG_{ij} = \frac{1}{D} \sum_{d=1}^{D} \sqrt{\frac{1}{t} \sum_{e=j-t}^{j} (S_{ie} - \mu_i)^2}
\end{equation}

where $\mu_i = \frac{1}{t} \sum_{e=j-t}^{j} S_{ie}$ and $t$ is the number of previous epochs used to compute the variance. If $t=5$, VOG can be computed only after the $5^{th}$ epoch.\\

Co-teaching \cite{han2018co} is a loss-based selection approach that separates clean samples from noisy labels as $Cl = \{b \in B \mid \mathcal{L}(b)$ $\text{ are the }$ $R \text{ smallest values}\}$, where $B$ is the mini-batch and $\mathcal{L}$ is the loss value, and $R$ is the number of examples to be selected as clean given by $R = \lfloor(1- \tau)*B\rfloor$. Usually, the forget rate \(\tau\) is chosen to match the noise rate \(p\). In our approach, i.e, Co-teaching VOG, we select clean samples as: $Cl = \{b_1 \in B \mid \mathcal{L}(b_1) \text{ are the } R_1 \text{ smallest values}\}$ $\cup$ $\{b_2 \in B \setminus b_1 \mid VOG (b_2) \text{ are the } R_2 \text{ smallest values}\}$,  where $R_1 = \lfloor(1-m)*R\rfloor$, $R_2 = \lfloor m * R\rfloor$, and $m$ is a hyperparameter we refer to as mix ratio. When $m =0$, no examples are selected using VOG. We only employ the VOG after a warm-up phase, because VOG is unstable in the early phase. The noisy subset is given by $\hat{Cl}$ = $B \setminus Cl$. At the end of the training, we combine the samples selected at each mini-batch to obtain all the clean and noisy samples from the entire dataset containing \(N\) samples. Let $\hat{\mathbf{Cl}}$ represent the noisy samples set and $\mathbf{Cl}$ represent the clean samples set, from the whole dataset.

\subsection{Active Label Cleaning}

The first stage identifies $\mathbf{Cl}$ clean samples, while the remaining noisy samples $\hat{\mathbf{Cl}}$ undergo a label correction in this phase. We have a predefined annotation budget $A_l$ that denotes the number of examples that we can afford to relabel. The noisy samples are annotated in batches up to $M$ annotation rounds, at the rate of $a_l$ samples per round. We apply an active learning sampler to select the most important samples which, after label correction, would improve test performance with fewer annotation rounds, given by: $L = {\mathrm{argmax}}_{_{{L \subseteq \hat{\mathbf{Cl}}, |L| = a_l}}} \mathbf{\Phi}(x | x \in \hat{\mathbf{Cl}})$, where $\mathbf{\Phi}$ is the scoring function. We then pass the selected samples to an expert annotator for relabeling: $L_{clean} = \mathcal{ORACLE}(L)$. After cleaning $L$ samples, we update the noisy set and the clean set as $\hat{\mathbf{Cl}} : \hat{\mathbf{Cl}} \setminus L_{clean}$, and $\mathbf{Cl} : \mathbf{Cl} \cup L_{clean}$, respectively. After each annotation round, the original model is retrained with the updated clean set.

\section{Experiments}

\subsection{Datasets}
\begin{itemize}[label={},leftmargin=*]
    \item \textbf{Long-tailed NCT-CRC-HE-100K:} We created a long-tailed dataset from the original NCT-CRC-HE-100K \cite{kather2019predicting} by modifying the class distribution. The original dataset has 100,000 histopathology images for training and 7,180 for testing, with nine classes. To create a long-tailed version, we randomly sampled examples from the training set using the Pareto distribution \cite{cui2019class}: \(N_c = N_0(r^{-(k-1)})^c\), where \(k\) is the total number of classes and \(c\) is the class being sampled. Here, we chose \(N_0 = \min(N_0, N_1, \ldots, N_{k-1})\), representing the class with the minimum number of samples in the original dataset, and \(r = 100\) as the imbalance factor. After creating the imbalanced dataset, we divided the training set into training and validation sets with split ratios of 0.8 and 0.2, respectively. Consequently, the final long-tailed training set contains 15,924 samples, and the validation set contains 3,982 samples, while the original test set remains unchanged. Label noise is injected only in the training set.\\
    
    \item \textbf{ISIC-2019:} ISIC-2019\footnote{https://challenge.isic-archive.com/landing/2019/} is an imbalanced dataset, which comprises 25,331 RGB images, each belonging to one of eight skin disease conditions. We divided the original dataset into training, validation, and test sets randomly, using split ratios of 0.7, 0.1, and 0.2, respectively. As a result, the training, validation, and test sets contain 17,731, 2,533, and 5,067 samples, respectively, where label noise is only injected into the training set.
\end{itemize}

\subsection{Baselines}
We compared our proposed method -- Co-teaching VOG ($\text{CT}_\text{VOG}$) + Active Learning (Random, Entropy, Coreset \cite{sener2018active}), where we robustly train the model using Co-teaching, regularized by VOG in sample selection -- against the following baselines:  1) Active learning (Random and Entropy \cite{cohn1996active}), where we directly clean a few samples, exhausting annotation budget $a_l$, and train the initial base model. Then, we gradually clean additional samples, selected by active learning at each round, and fine-tune the model. This approach does not involve training with noisy labels in the initial phase. 2) Cross-Entropy (CE) + Active Learning (Random, Entropy), where we initially train the model using the noisy dataset, then gradually clean the samples selected by AL and fine-tune the model using only cleaned data. 3) ALC w/ Co-teaching (Bernhardt {\it et al.} \cite{bernhardt2022active}), where the model was fine-tuned using Co-teaching in each round using available data (both cleaned and still noisy). At each round, it uses a ranker function to select the samples to be cleaned. Since this method is primarily proposed for multi-annotator settings, we adapted it to our single-annotator setting and implemented it. 


In our method, the samples selected in the initial phase as clean are retained and used to train the classifier using standard cross-entropy loss. The remaining noisy samples selected via active learning are gradually cleaned and added to this clean set at each round, while simultaneously tuning the model. \textit{Evaluation: Since all our datasets are imbalanced, we evaluate the performance of our method and all the baseline performance using the macro-averaged F1-score, which captures both precision and recall, in the test set.} The test score is computed in the epoch where the validation set performed the best.


\subsection{Implementation Details}

We used ResNet18 pretrained on ImageNet as the feature extractor backbone for all our experiments. The batch size (\(b\)) was set to 256, and we trained the model with the SGD optimizer, an initial learning rate of 0.01, a momentum of 0.9, weight decay of \(1 \times 10^{-4}\), and cosine scheduler. These parameters remained consistent across both datasets. Images from both datasets were resized to \(244 \times 224\), and we applied basic data augmentations, including random crop, random flip, random Gaussian blur, and random color jittering. We selected two high noise rates, \(p = \{0.4, 0.5\}\) and \(p = \{0.7, 0.8\}\), for ISIC-2019 and Long-tailed NCT-CRC-HE-100K, respectively, the rates at which the classification performance degradation is high.

For Co-teaching VOG, we set the warm-up epoch to 10. The number of instances selected as clean is determined by the forget rate, which depends on the maximum forget threshold \(\tau\) and the decay rate \(c\). Following \cite{han2018co}, we set \(\tau = p\) and \(c = 1\), where \(p\) represents the label noise rate. The mix ratio (\(m\)) for Co-teaching VOG is a hyperparameter that depends on the dataset and noise rate. We found \(m = 0.2\) and \(m = 0\) to work best for ISIC-2019 at \(p = 0.5\) and \(p = 0.4\), respectively. Similarly, \(m = 1\) was optimal for Long-tailed NCT-CRC-HE-100K for both \(p = 0.8\) and \(p = 0.7\).

In the active label cleaning phase, we set the annotation round (\(M\)) to 8. The per-round annotation budget (\(a_l\)) varied for both datasets and label noise rates. In ISIC-2019, \(a_l = \{394, 492\}\) for \(p = \{0.4, 0.5\}\), and in Long-tailed NCT-CRC-HE-100K, \(a_l = \{70, 80\}\) for \(p = \{0.7, 0.8\}\). We ran experiments across three seeds to obtain the average and standard deviation. All the training sessions were performed using the PyTorch 1.12.1 framework in Python 3.9 on a single A100 GPU.

\section{Results}
\subsection{Overall Active Label Cleaning}

\begin{figure}[!ht]
    \centering
    \footnotesize
    \begin{subfigure}[b]{0.49\textwidth} 
        \centering
        \includegraphics[width=\textwidth]{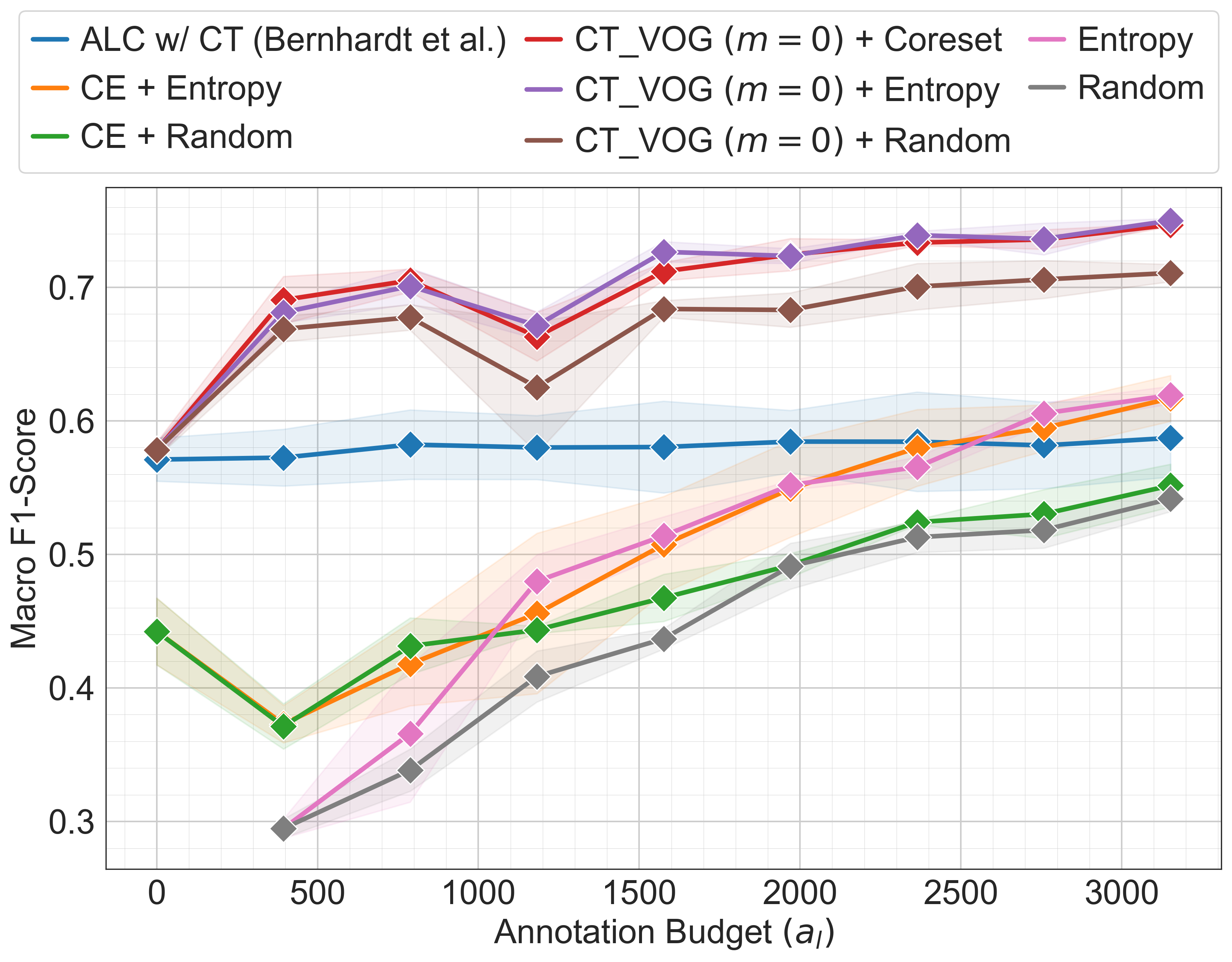}
    \end{subfigure}
    \begin{subfigure}[b]{0.50\textwidth} 
        \centering
        \includegraphics[width=\textwidth]{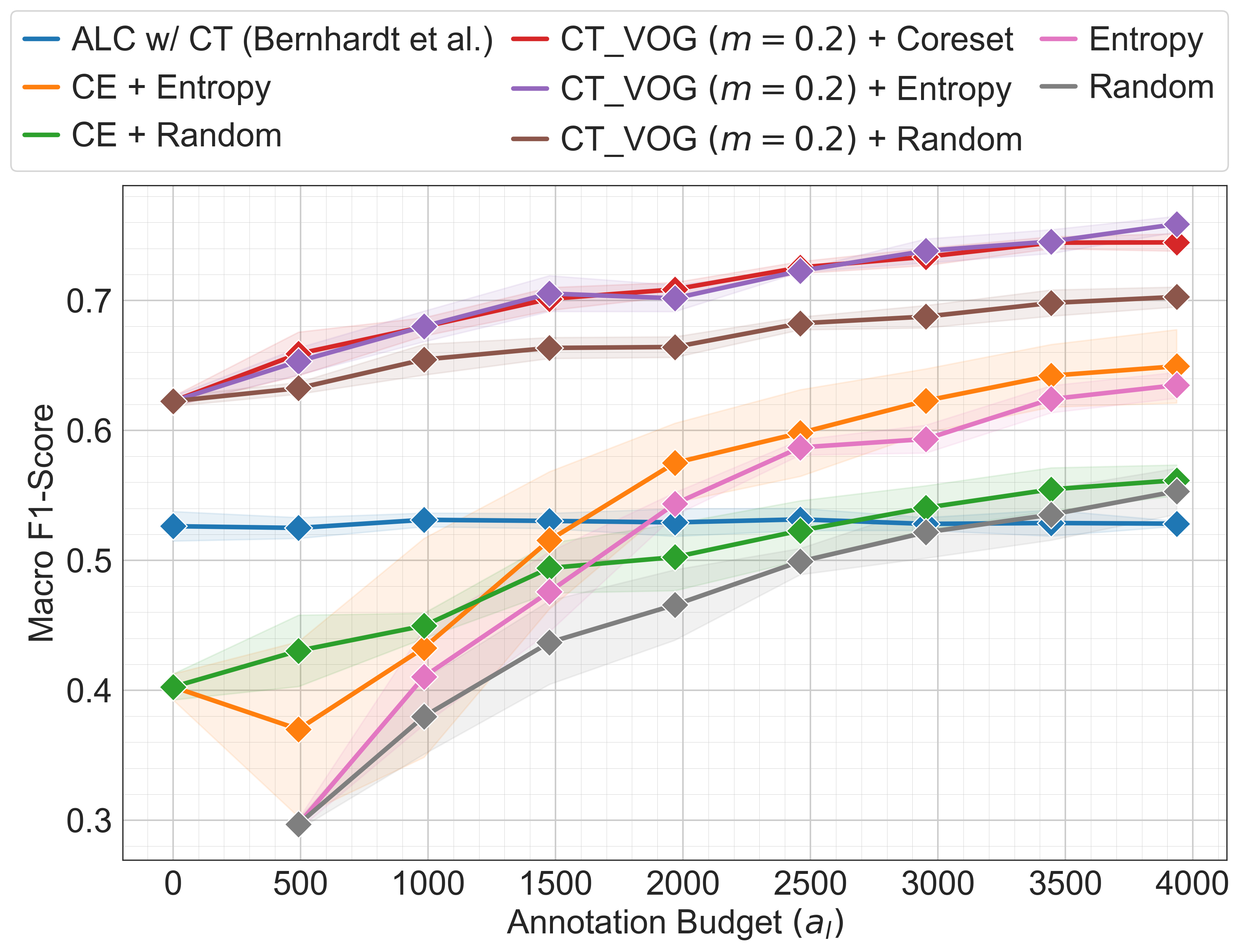}
    \end{subfigure}
    \caption{Comparison of the macro-averaged test F1-score across various baselines in ISIC-2019 dataset at two noise rates : $p = 0.4$ (left) and $ p =0.5$ (right).}
    \label{fig:active_label_cleaning_ISIC}
\end{figure}

\begin{figure}[!ht]
    \centering
    \begin{subfigure}[b]{0.49\textwidth} 
        \centering
        \includegraphics[width=\textwidth]{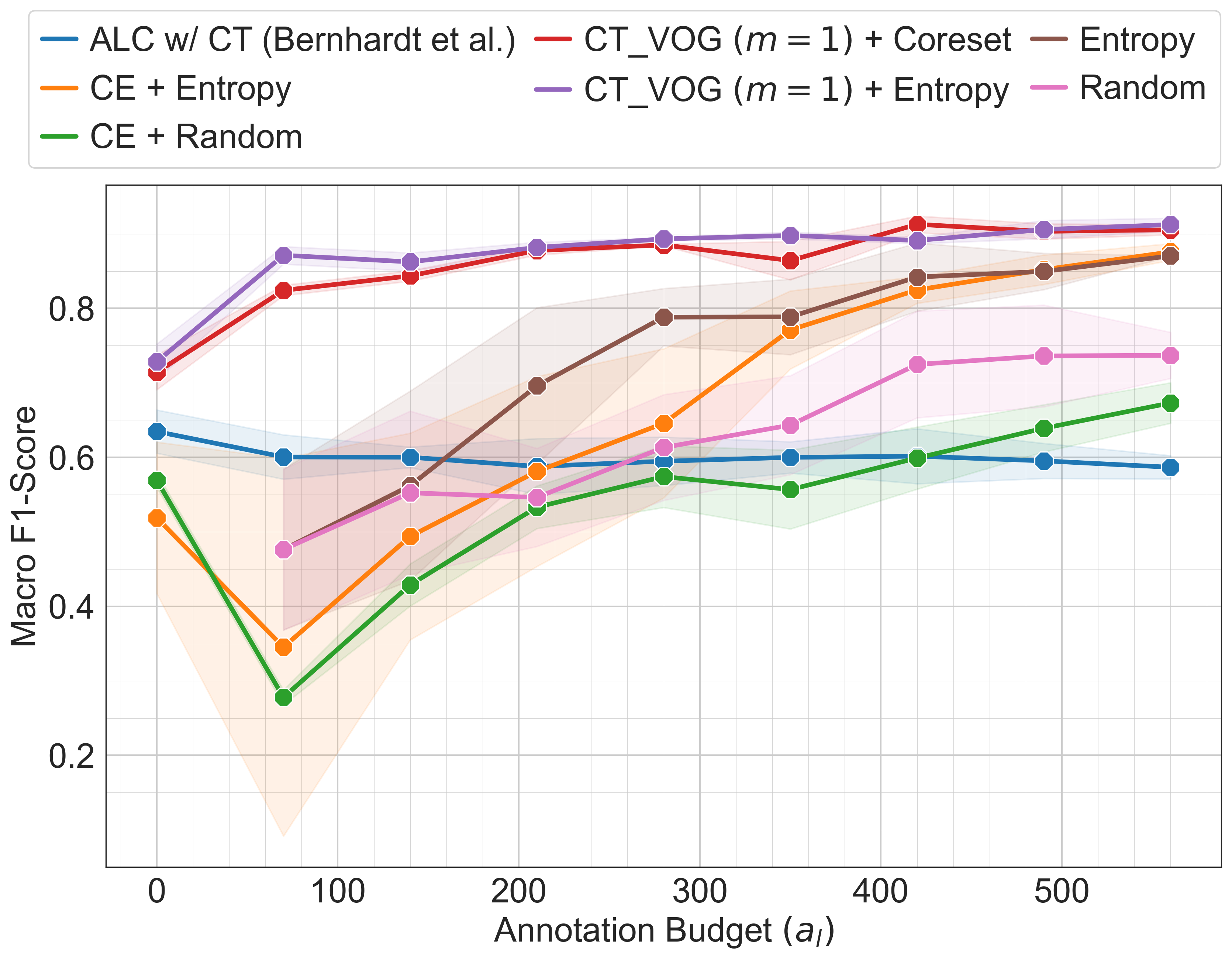}
    \end{subfigure}
    \begin{subfigure}[b]{0.49\textwidth} 
        \centering
        \includegraphics[width=\textwidth]{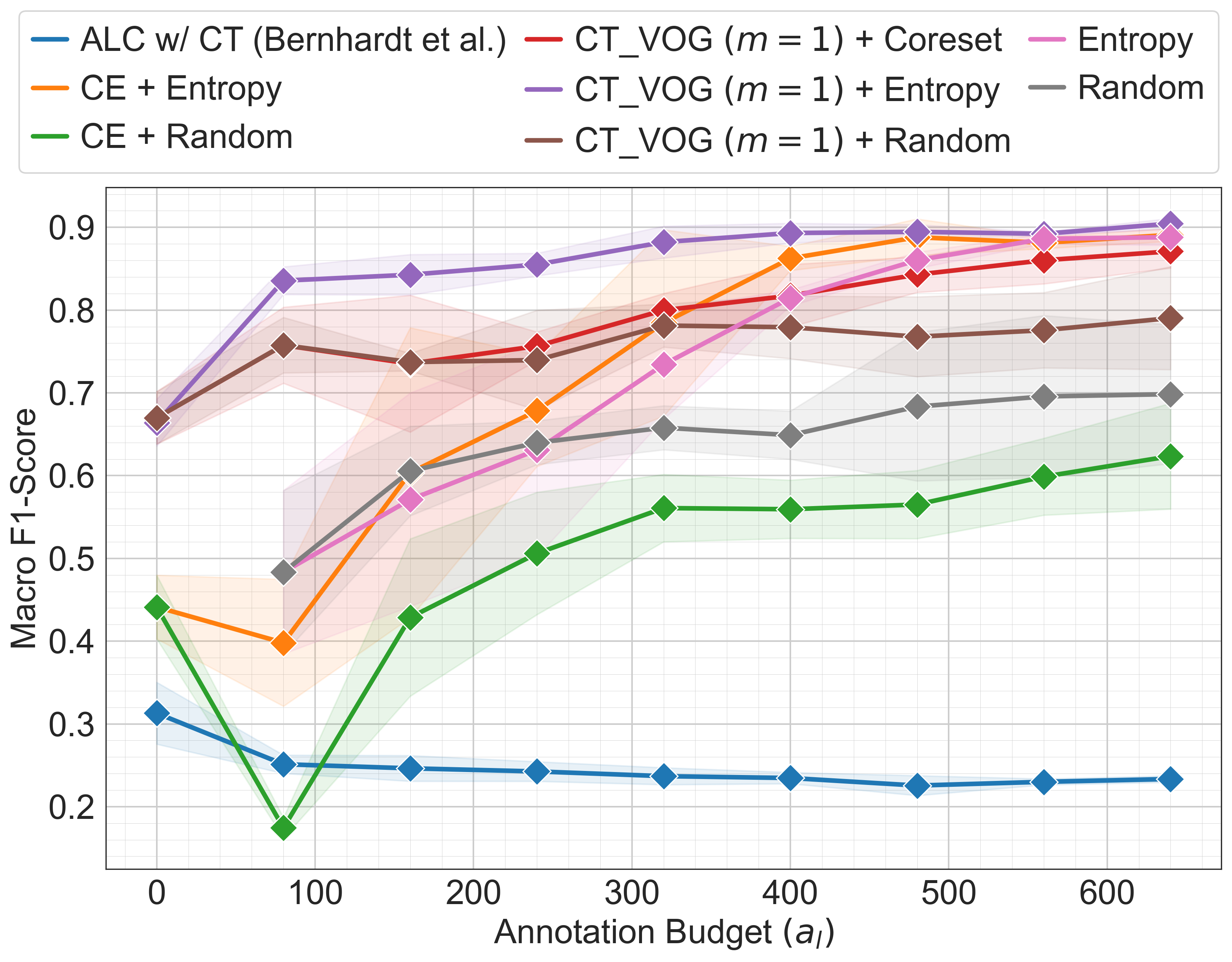}
    \end{subfigure}
    \caption{Comparison of the macro-averaged test F1-score across various baselines in Long-tailed NCT-CRC-HE-100K dataset at two noise rates: $p = 0.7$ (left) and $ p =0.8$ (right).}
    \label{fig:active_label_cleaning_LT_Histo}
\end{figure}

In Fig. \ref{fig:active_label_cleaning_ISIC} and Fig. \ref{fig:active_label_cleaning_LT_Histo}, we benchmark the performance of our approach against various baseline methods on the ISIC-2019 and the long-tailed NCT-CRC-HE-100K datasets, respectively. Our LNL strategy ($\text{CT}_\text{VOG}$) delivers a substantial performance uplift from the beginning, prior to any label cleaning efforts. Then, the active learning strategies, Entropy and Coreset, effectively select important noisy samples to re-label. These samples further enhanced the model's performance. Since strategies that rely solely on active learning (Random/Entropy) initially require clean samples to train the model, they consequently exhaust a round budget from the start. Initially, training with noisy labels using standard cross-entropy (CE) proves more advantageous than just solely relying on active learning from the beginning. However, the performances converge to solely using active learning in the later stages, with additional label-cleaning rounds. We observed no further improvement upon cleaning additional labels when using ALC with Co-teaching (Bernhardt {\it et al.} \cite{bernhardt2022active}). This method, proposed for multi-annotator settings and not intended for imbalanced datasets, still selected the same initial examples as clean, even after the noisy labels had been cleaned.

It is important to note that the macro-averaged test F1-score after training on the ISIC-2019 dataset with entirely clean labels is \(0.767 \pm 0.004\). Our method achieves this performance by relabeling merely 3,152 samples at a noise rate of 0.4 and 3,936 samples at a noise rate of 0.5, out of 17,731 training examples. Similarly, the macro-averaged test F1-score for the long-tailed NCT-CRC-HE-100K dataset is \(0.894 \pm 0.12\) with all clean labels. Our approach matches this score by relabeling just 300 samples at a noise rate of 0.7 and 400 samples at a noise rate of 0.8, from a total of 15,924 available training samples.

\subsection{VOG as a regularizer for LNL}

\begin{table}[!ht]
\centering
\caption{Comparing LNL performance of Co-teaching (CT) alone vs. Co-teaching with VOG as a regularizer (\(\text{CT}_{VOG}\))  on the ISIC-2019 dataset with a label noise rate (\(p=0.5\)). We examine the Recall and Guess \%, which indicate the proportion of samples identified as clean that belong to underrepresented classes.}
\label{tab:VOG_reg}
\begin{tabular}{l|c|c|c|c|c|c}
\hline
& \multicolumn{2}{c|}{DF} & \multicolumn{2}{c|}{VASC} & \multicolumn{2}{c}{SCC} \\ \hline
LNL  & Recall & Guess (\%) & Recall & Guess (\%) & Recall & Guess (\%) \\
\hline
CT                 & 0.00$\pm$0.00 & 0.00$\pm$0.00 & 0.39$\pm$0.34 & 50.73$\pm$43.94 & 0.26$\pm$0.22 & 36.11$\pm$31.30 \\
$\text{CT}_{VOG}$ & 0.10$\pm$0.09 & 20.00$\pm$17.33 & 0.57$\pm$0.00 & 80.44$\pm$3.92 & 0.37$\pm$0.02 & 52.38$\pm$2.06 \\
\hline
\end{tabular}
\end{table}

In Table \ref{tab:VOG_reg}, we investigate the benefits of integrating VOG into Co-teaching ($\text{CT}_{VOG}$) for enhancing the identification of underrepresented samples. In ISIC-2019 at a noise rate of $p=0.5$, we observed that Co-teaching alone tends to overlook minority classes while identifying clean samples (see class DF). By regularizing the sample selection with VOG, the accuracy of identifying samples from underrepresented classes improves, resulting in enhanced performance in LNL at the initial phase. 

\section{Discussion and Conclusion}

In this work, we present a strategy that combines learning with noisy labels and active learning to actively relabel noisy samples, thereby enhancing medical image classification performance in the presence of noisy labels. Our method of regularizing Co-teaching with VOG for sample selection has proven to handle imbalanced cases better. We show that by relabeling only a few samples, our method can match the performance achieved with clean labels in the ISIC-2019 and long-tailed NCT-CRC-HE-100K datasets. 

While our method shows promising results compared to the baseline, some limitations in our work could be addressed in future research. First, we limited our study to a single CNN-based model. Exploring the behavior of larger models would be an interesting extension. Additionally, we adhered to common standard protocols by limiting our study to a uniform label noise distribution and specific noise rates. Investigating the performance of our method under different types of noise would provide further insights. Finally, it would also be valuable to examine how our method performs across various levels of skewness in imbalanced distributions.

\begin{credits}
\subsubsection{\ackname} Research reported in this publication was supported by the NIGMS Award No. R35GM128877 of the National Institutes of Health, and by OAC Award No. 1808530 and CBET Award No. 2245152, both of the National Science Foundation, and by the Aberdeen Startup Grant CF10834-10. We also acknowledge Research Computing at the Rochester Institute of Technology \cite{RITRC} for providing computing resources. We would also like to thank Dr. Bishesh Khanal from NAAMII for his suggestions during the early discussions.

\end{credits}

%
%
\bibliographystyle{splncs04}
\bibliography{Paper-3963}
\input{Supplementary}
\end{document}

%% file: Supplementary.tex
\chapter*{Supplementary Materials}

\section{Detailed Dataset Information}

\begin{figure}[h!]
\centering
\includegraphics[width=1\linewidth]{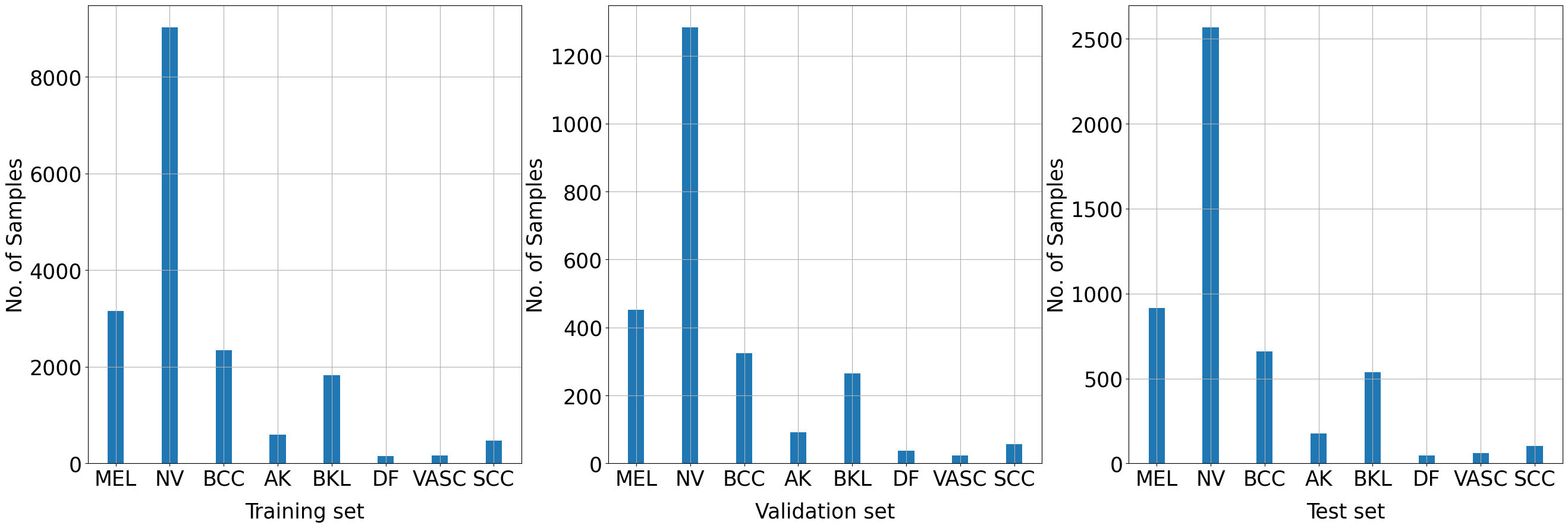}
\caption{Distribution of the classes in ISIC-2019 Dataset}
\label{fig:isic_distribution}
\end{figure}

\begin{figure}[h!]
\centering
\includegraphics[width=1\linewidth]{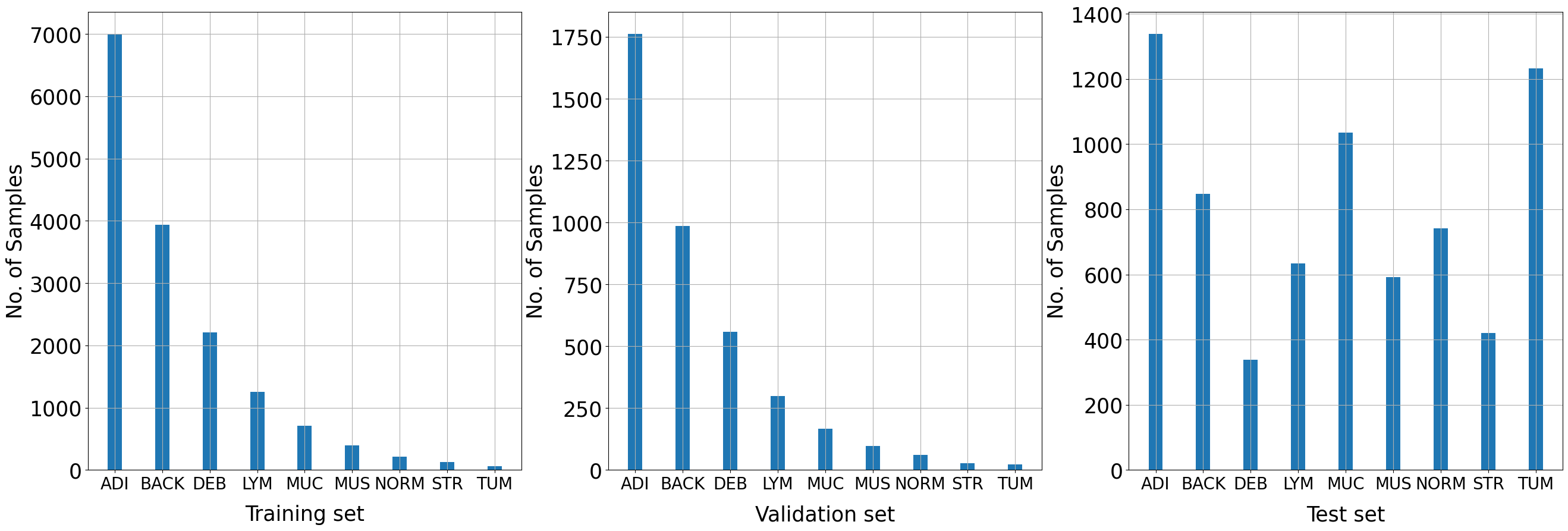}
\caption{Distribution of the classes in long-tailed NCT-CRC-HE-100K Dataset}
\label{fig:LT_histopathology_distribution}
\end{figure}

\begin{figure}[ht!]
    \centering
    \begin{subfigure}[b]{1\textwidth} 
        \centering
        \includegraphics[width=\textwidth]{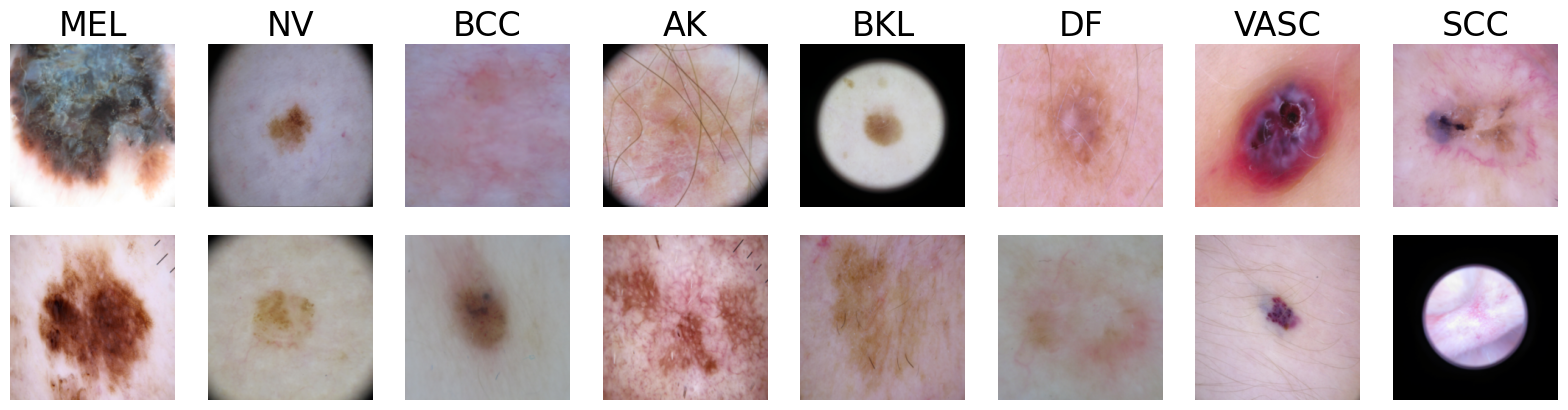}
        \caption{ISIC-2019}
    \end{subfigure}
    \begin{subfigure}[b]{1\textwidth} 
        \centering
        \includegraphics[width=\textwidth]{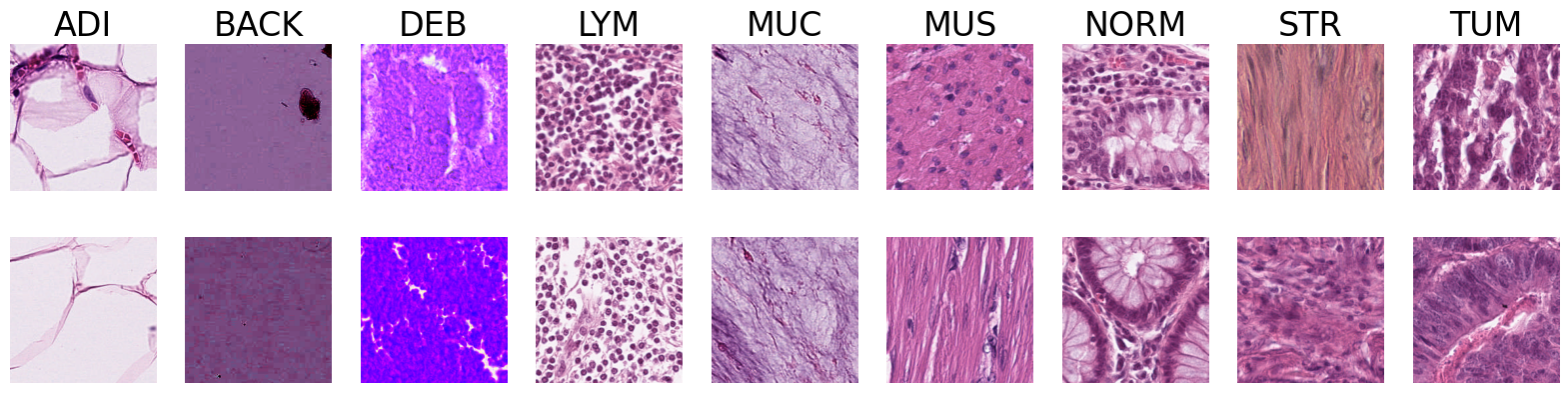}
        \caption{Long-tailed NCT-CRC-HE-100K}
    \end{subfigure}
    \caption{Some examples of images of the two datasets}
    \label{fig:sample_images}
\end{figure}

We illustrate the class distribution across each dataset in Fig. \ref{fig:isic_distribution} and Fig. \ref{fig:LT_histopathology_distribution}, highlighting the significant imbalance. Additionally, we present representative samples from each class in Fig. \ref{fig:sample_images}.

\section{Hyperparameter Analysis: Mix Ratio ($m$)}
\begin{figure}[ht!]
    \centering
    \begin{subfigure}[b]{0.49\textwidth} 
        \centering
        \includegraphics[width=\textwidth]{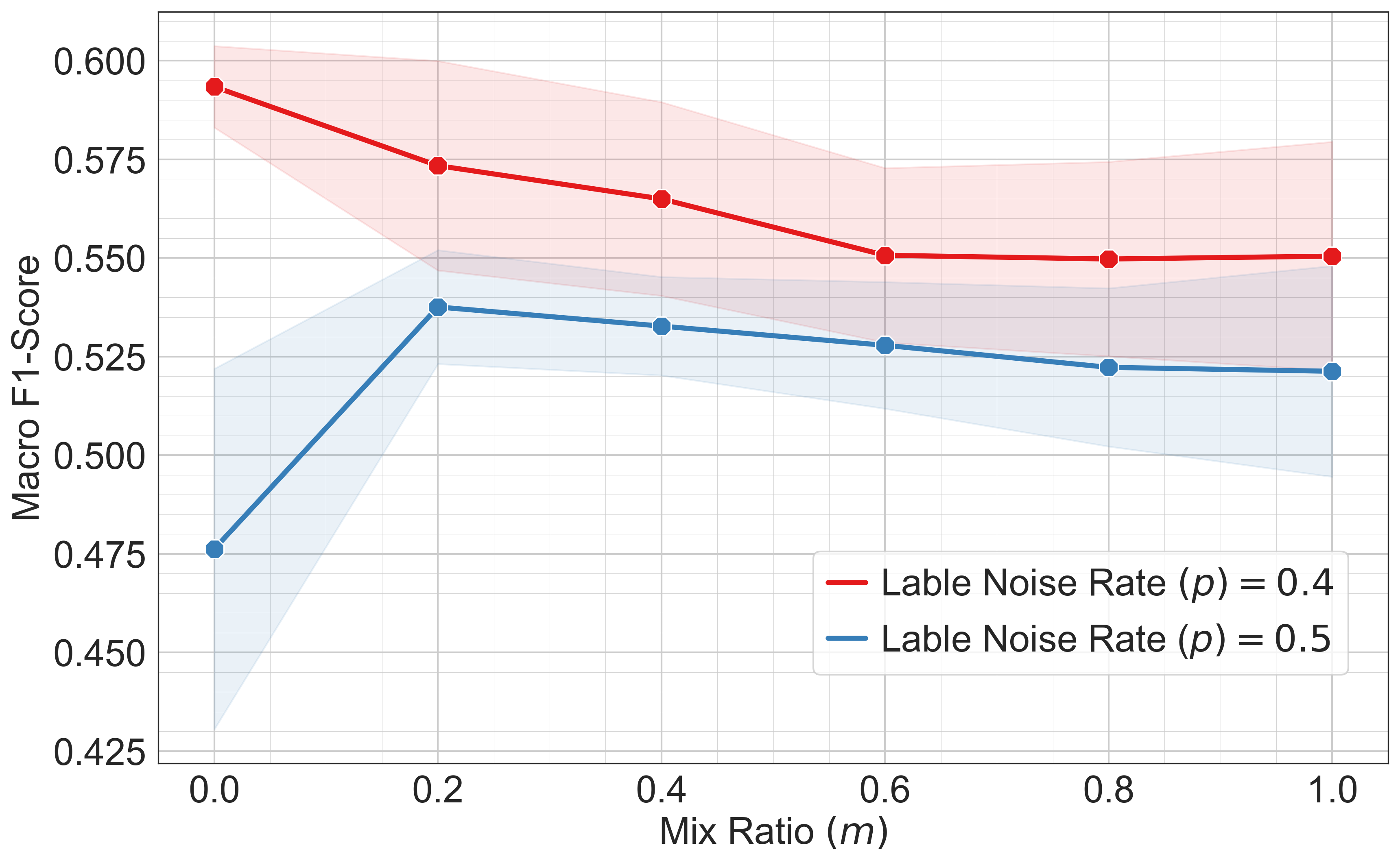}
        \caption{ISIC-2019}
    \end{subfigure}
    \begin{subfigure}[b]{0.49\textwidth} 
        \centering
        \includegraphics[width=\textwidth]{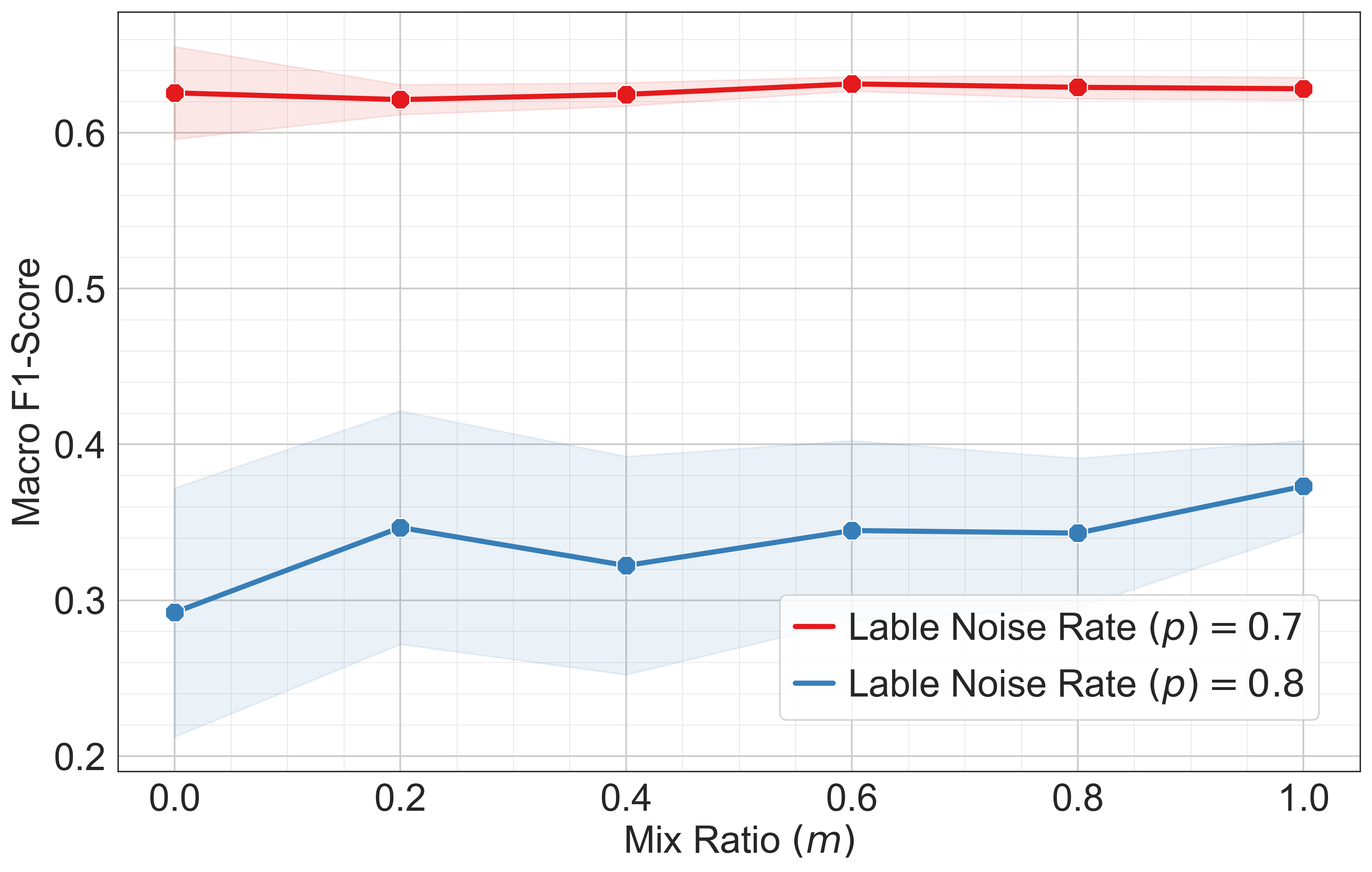}
        \caption{LT NCT-CRC-HE-100K}
    \end{subfigure}
    \caption{Hyperparemeter study of mix ratio ($m$) in two datasets, when training Co-teaching VOG at the first phase.}
    \label{fig:hyper_mix_up}
\end{figure}

In Fig. \ref{fig:hyper_mix_up}, we compare the impact of the mix ratio ($m$) in Co-teaching VOG, using the macro-averaged test F1-score obtained after training with noisy labels in the initial LNL phase. These results indicate that this hyperparameter differs across datasets and can vary with label noise ($p$).

\section{Base Model Initialization Strategy}

\begin{figure}[ht!]
    \centering
    \begin{subfigure}[b]{0.49\textwidth} 
        \centering
        \includegraphics[width=\textwidth]{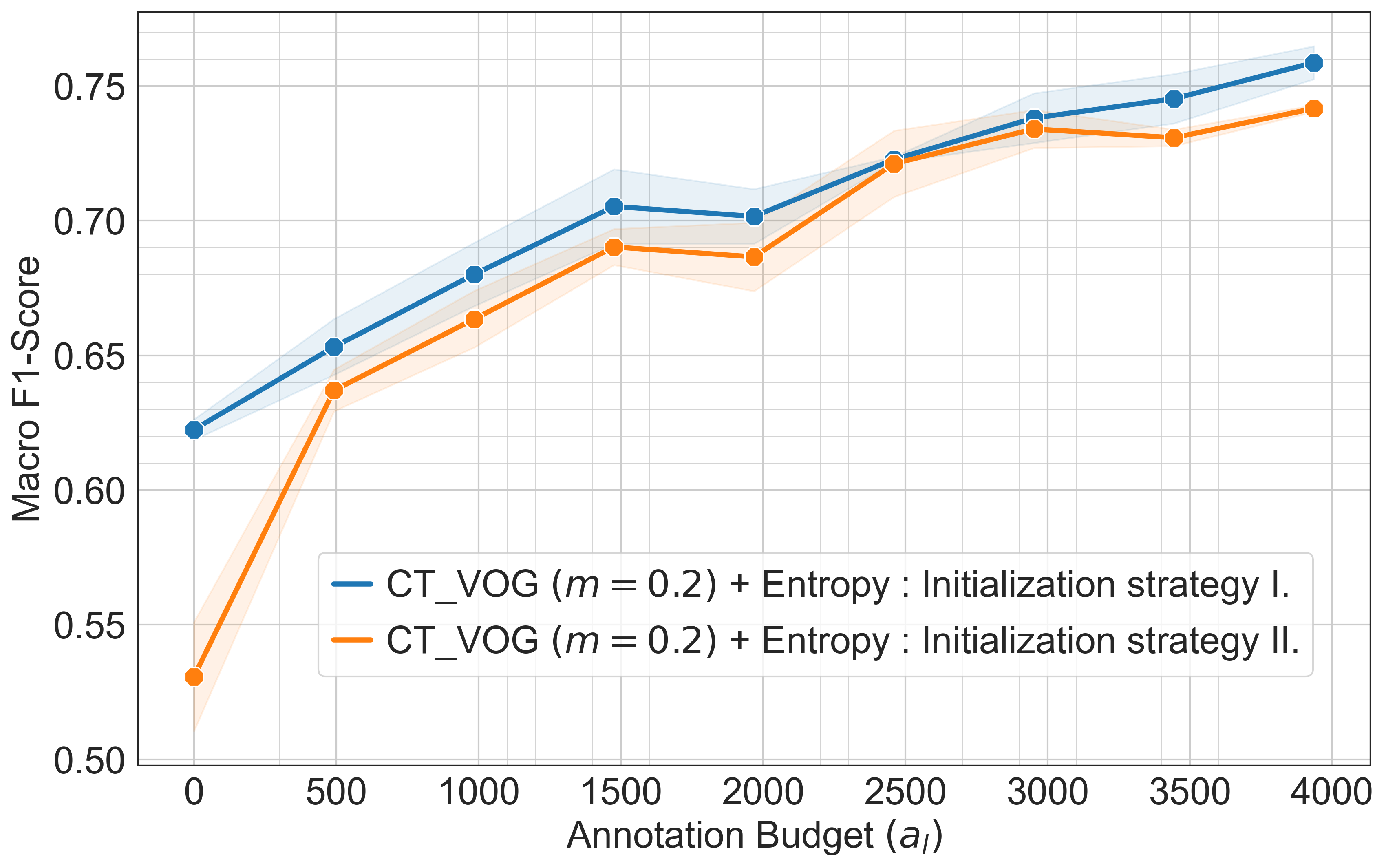}
        \caption{ISIC-2019 ($p = 0.5$)}
    \end{subfigure}
    \begin{subfigure}[b]{0.49\textwidth} 
        \centering
        \includegraphics[width=\textwidth]{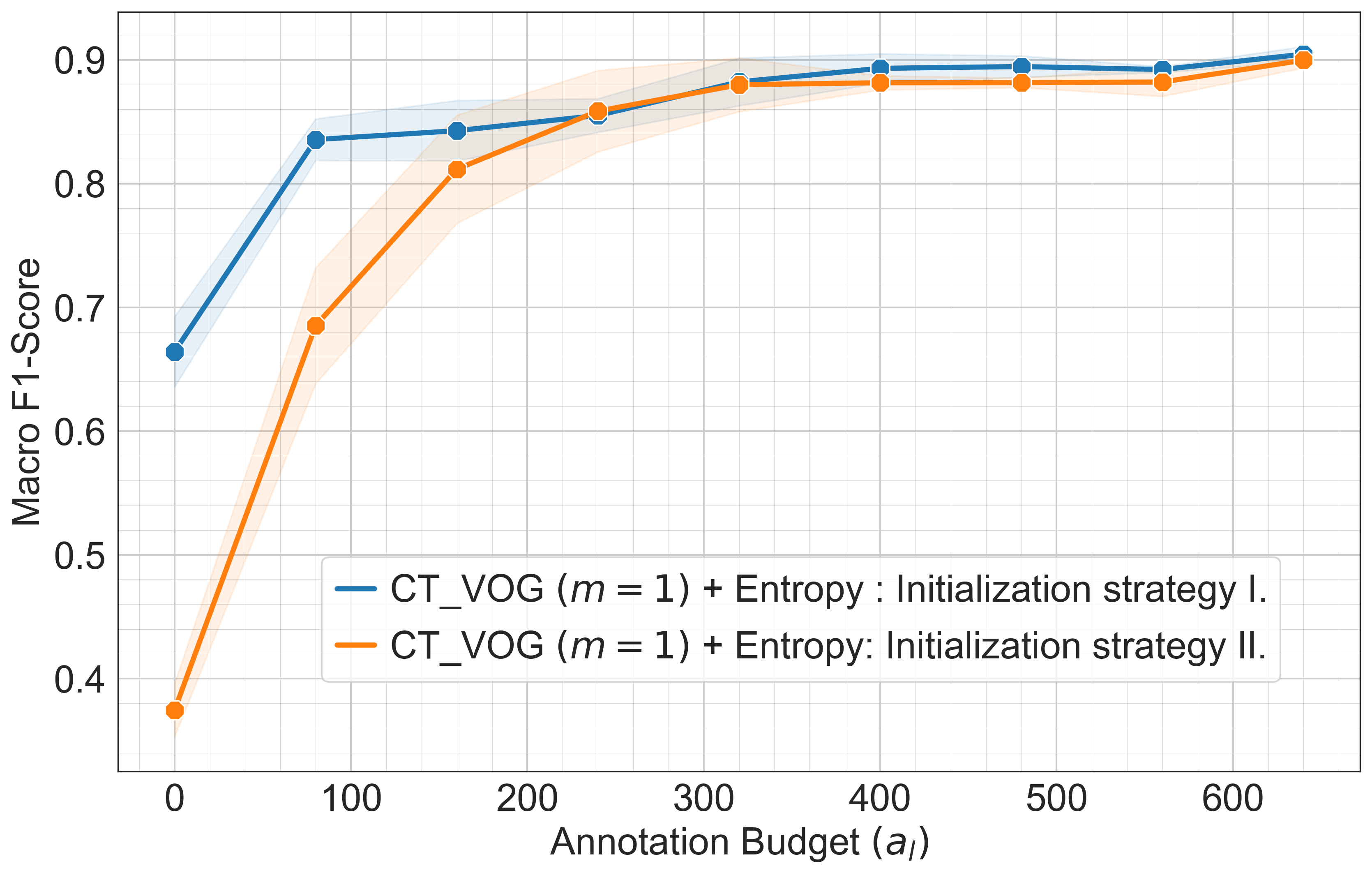}
        \caption{LT NCT-CRC-HE-100K ($p = 0.8$)}
    \end{subfigure}
    \caption{Comparison of test performance (macro-averaged F1-score) using two initialization strategies across two label noise rates (\(p\)) in two datasets. Initialization strategy I. refer to retraining the model from scratch using cross-entropy with only the clean labels identified by Co-teaching VOG. Initialization strategy II. refers to directly utilizing the model trained with Co-teaching VOG on noisy labels.}
    \label{fig:initializing_strategy}
\end{figure}

There are two strategies to initialize the base model in the first phase before the active label cleaning round begins: I. either use the model trained on a noisy dataset using Co-teaching VOG (similar to \cite{bernhardt2022active}) or II. use the samples selected by Co-teaching VOG as clean labels to train a new model using standard cross-entropy loss. In Fig. \ref{fig:initializing_strategy}, we compared these strategies and observed that separately training the model using standard cross-entropy with only the samples identified by Co-teaching VOG as clean labels improved the initial performance the most. We argue that by segregating the noisy samples from an early stage, we reduce the possibility of model distortion due to noisy labels. Therefore, we adopted strategy I. for Co-teaching VOG, as reported in the Results section.

